\pdfoutput=1

\documentclass[11pt]{article}

\usepackage[]{emnlp2021}

\usepackage{times}
\usepackage{latexsym}
\usepackage{times}
\usepackage{latexsym}

\usepackage{graphicx}
\usepackage{todonotes}
\usepackage{multirow}
\usepackage{enumitem}

\usepackage[T1]{fontenc}

\usepackage[utf8]{inputenc}

\usepackage{microtype}

\title{ConQX: Semantic Expansion of Spoken Queries for Intent Detection based on Conditioned Text Generation}

\author{Eyup Halit Yilmaz \\
  Aselsan Research Center \\
  Ankara, Turkey \\
  \texttt{ehyilmaz@aselsan.com.tr} \\\And
  Cagri Toraman \\
  Aselsan Research Center \\
  Ankara, Turkey \\
  \texttt{ctoraman@aselsan.com.tr} \\}

\begin{document}
\maketitle
\begin{abstract}
Intent detection of spoken queries is a challenging task due to their noisy structure and short length. To provide additional information regarding the query and enhance the performance of intent detection, we propose a method for semantic expansion of spoken queries, called ConQX, which utilizes the text generation ability of an auto-regressive language model, GPT-2. To avoid off-topic text generation, we condition the input query to a structured context with prompt mining. We then apply zero-shot, one-shot, and few-shot learning. We lastly use the expanded queries to fine-tune BERT and RoBERTa for intent detection. The experimental results show that the performance of intent detection can be improved by our semantic expansion method.
\end{abstract}

\section{Introduction}
In human-to-machine conversational agents, such as Amazon Alexa and Google Home, \emph{intent detection} aims to identify user intents that determine the command to be executed. \emph{Spoken query}, also called as \emph{utterance}, can be classified into a set of pre-defined user intents \cite{Tur:2010}. 

Intent detection is a challenging task due to the noisy, informal, and limited structure of spoken queries. Detection models may suffer from the problems of sparsity, ambiguity, and limited vocabulary. Recent state-of-the-art language models, such as GPT-2 \cite{Radford:2019} and GPT-3 \cite{Brown:2020} based on the Transformer architecture \cite{Vaswani:2017}, incorporate domain-independent large corpora in training. They have the capability of coherent text generation when the task is prompted in natural language. The clarification of short spoken queries can be done by generating coherent and semantically related text. For instance, the given query ``what is amzn worth" is expanded to ``what is amzn worth \emph{what is Amazon's stock worth}" that clarifies \emph{stock worth}, as well as solving the ambiguity in \emph{amzn}.

Transformer-based text generation does not always produce meaningful text segments \cite{shao-etal-2017-generating}. For instance, the given query without conditioning ``has my card application processed yet?" is expanded to ``has my card application processed yet? \emph{If you are not yet with us}" that gets a trivial text segment given in italic. The reason would be that the model does not know the context of card application, such as banking or membership card. However, the input can be conditioned with a better prompt ``[I am a bank customer], has my card application processed yet?" that gets additional context as bank customer. The input would then be expanded to a non-trivial text segment in the context of bank cards. 

In order to solve the problems regarding the noisy and limited structure of spoken queries, we propose a novel method, called ConQX, for semantic expansion of spoken queries with conditioned text generation. The method name refers to \textbf{Con}ditioned spoken \textbf{Q}uery e\textbf{X}pansion. Specifically, we employ a Transformer-based language model, namely GPT-2, to generate semantically related text segments. We condition the input query to set up a structured context for generating text segments. For conditioning, we mine useful prompts that provides structured context, as we call \emph{prompt mining}. We then append the generated text segments to existing spoken queries, and fine-tune state-of-the-art language models, namely BERT \cite{Devlin:2018} and RoBERTa \cite{Liu:2019b}, for the downstream task of intent detection. 

Conditioned expansion aims to describe the task to the model in natural language and provide a number of ground truth demonstrations of the task at inference time. To exploit conditioned expansion, we examine zero-shot, one-shot, and few-shot learning \cite{Brown:2020}.

Traditional semantic expansion methods rely on keyword-based expansion, which utilizes proximity in a semantic space regardless of contextual coherence \cite{Roy:2016}. However, the models using contextual word embeddings, such as BERT, are shown to benefit from natural language queries that keep the grammar structure and word relations \cite{Padaki:2020,Dai:2019}. Transformer-based text generation can output more coherent natural language queries, compared to keyword-based expansion \cite{Radford:2019} that mostly adapts to improve the performance of retrieval algorithms \cite{Claveau:2020}.

The language model can be adapted to the downstream task with natural language prompts to achieve competitive performance. The design of an input prompt is important, since different writings of the task can affect the performance significantly \cite{jiang-etal-2020-know}. Although there are some efforts for the automation and standardization of prompt generation \cite{jiang-etal-2020-know, Gao:2020}, they do not consider long text generation tasks, as in the case of semantic expansion. We employ prompt mining on a set of manually generated conditioning prompts and experiment with zero-shot, one-shot, and few-shot learning to further adapt the language model to the task of semantic expansion.

\section{Conditioned Query Expansion}
Given a set of spoken queries and input prompts, we use a pre-trained Transformer-based language model, GPT-2 \cite{Radford:2019}, to generate coherent text.
A spoken query is placed in manually generated natural language prompts that are determined with prompt mining, and given as input to GPT-2 with zero/one/few-shot learning. The generated text segment is appended to the end of the original query to obtain the expanded query. BERT \cite{Devlin:2018} and RoBERTa \cite{Liu:2019b} are then fine-tuned for the task of intent detection.

\subsection{Text Generation}
Our method for semantic expansion is based on language modeling \cite{Bengio:2003}, formulated in Equation \ref{equation:language_model}, where $q$ is a spoken query and $p(q)$ is the maximum likelihood probability of document estimation based on a sequence of tokens.

\begin{equation}
\label{equation:language_model}
    p(q) = \prod_{i=1}^{n} p(s_i | s_1, ..., s_{i-1})
\end{equation}

ConQX employs the conditional probability of estimating semantic expansion, $q'$, given the original query within the context of the input prompt, $p(q'|q)=p(s_{n+1},...,s_{n+k}|s_1,...,s_n)$, where $q$ has the length of $n$ tokens, and $q'$ has $k$ tokens. The likelihood is estimated by an auto-regressive language model that considers the distributions of previously generated tokens for next token prediction. 

\begin{figure*}[ht]
\centering
\frame{\includegraphics[scale=0.45, trim={-0.25cm 0cm -0.25cm -0.25cm}]{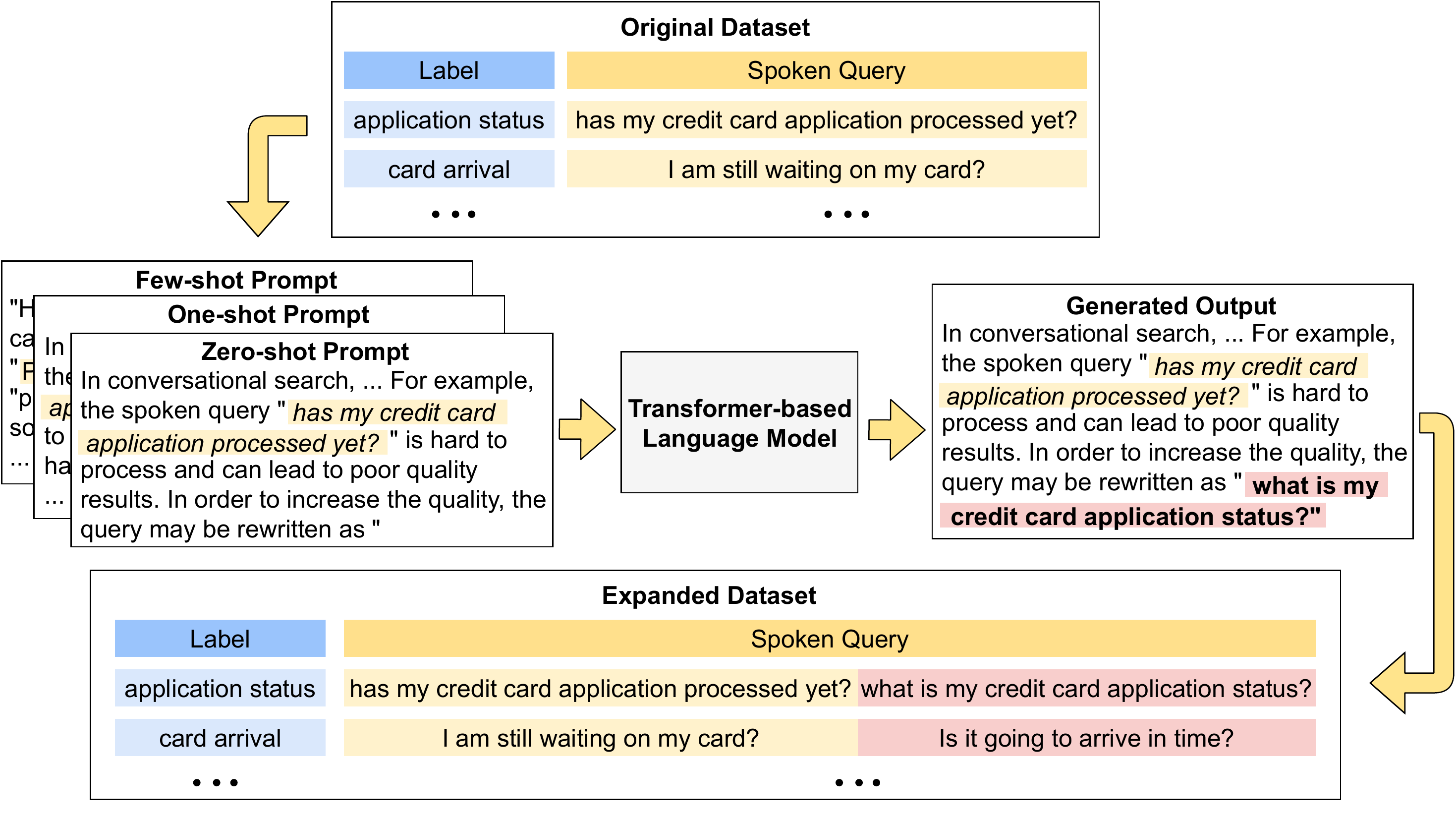}}
\caption{An example expansion process is illustrated for zero/one/few-shot learning. The input spoken query (labeled in yellow and italic) is inserted into quotations in a set of prompts. The generated text is then extracted to obtain the expanded query (labeled in red and bold). True demonstration(s) are also given in the prompts for one/few-shot learning.}
\label{InputPrompts}
\end{figure*} 

There are several methods to utilize auto-regressive text generation. Greedy search predicts the next token that has the highest probability of occurrence. However, greedy search does not generate coherent text due to repetitive results \cite{shao-etal-2017-generating}. We apply top-\emph{k} sampling that \cite{Fan:2018} predicts the next token from the most likely \emph{k} tokens to provide coherent and diverse text.

\subsection{Zero/One/Few-shot Learning}
Pre-trained language models are trained over large and domain-independent corpora. When used for a downstream task, such as semantic expansion in our case, they need conditioning to deduce the task and generate contextually related text. Zero-shot expansion aims to achieve this conditioning by inserting spoken queries into input prompts that contain natural language descriptions of the task without any demonstrations of the desired output. 

In one-shot expansion, the input prompt contains a ground-truth demonstration of the semantic expansion task. The language model is expected to infer the semantic expansion task more easily, compared to zero-shot expansion. Lastly, few-shot learning provides a number of true demonstrations to increase the performance of task inference.

Figure \ref{InputPrompts} shows an example spoken query for semantic expansion with zero/one/few-shot learning. The ability of task inference is known to be available in the large models in terms of the number of parameters, such as GPT-3 \cite{Brown:2020}; but also observed in smaller models, such as GPT-2 \cite{Schick:2020}. 

\subsection{Prompt Mining}
Determination of the proper input prompt for conditioning the language model can be achieved through prompt mining. The prompts provide additional context for the task inference of the language model. We manually generate a set of prompts that differ in text length, formality of the language, syntactic structure, and context. We apply empirical evaluation on the prompts, such that the classification performance is used to select a prompt after 10-fold leave-one-out cross-validation. 
We provide a subset of these prompts in Table \ref{AppendixInputPrompts}. 
\begin{table}[t]
\centering
\small
\caption{Prompt mining examples. [INP] is an input prompt, [EXP] is an expanded text segment.}
\label{AppendixInputPrompts}
\begin{tabular}{p{0.2cm}p{6.7cm}}
\hline
\textbf{} &\textbf{Input Prompt} \\
\hline
1 & [INP]. I would like to [EXP] \\ 
2 & Spoken queries are generally short and need to be expanded. For example, [INP] is hard to process and can lead to poor quality results. The query may be rewritten as ``[EXP]\\
3 & Voice Assistant: ``How can I help you?"
User: [INP]
Voice Assistant: ``Sorry, I didn't understand."
User: ``[EXP]\\
4 & In conversational search, spoken queries are short and need to be expanded. For example, [INP] is hard to process. The query may be rewritten as ``[EXP] \\
5 & I am a bank customer and I need support, [INP]. My intention is [EXP]\\
\hline
\end{tabular}
\end{table}

We provide a short prompt (the first), as well as a longer one (the second) that aims to exploit the ability of Transformer to model long-term dependencies with the Attention mechanism. The third prompt introduces a syntactic structure to condition the model, imitating a dialog. The fourth prompt is written in a more formal language, while the others in a daily language. The last one has additional context information as banking. Note that some of the prompts end with a quotation mark that enforces the language model to generate an example language; while the others generate expansions in the form of sentence completions.

\begin{table}[t]
\small
\centering
\caption{The details of the datasets used in this study.}\label{DataCharacteristics}
\begin{tabular}{lccc}
\hline
\textbf{Details}  & \textbf{Banking} & \textbf{CLINC} & \textbf{SNIPS}\\
\hline
Train samples & 10,003 & 18,000 & 13,084\\
Test samples & 3,080 & 4,500 & 700\\
Number of intents & 77 & 150 & 7\\
Avg. length (tokens) & 12.27 & 9.38 & 11.24\\
\hline
\end{tabular}
\end{table}

\section{Experiments}

\subsection{Datasets} 
We use three publicly available datasets for intent detection; namely Banking, CLINC, and SNIPS. Banking \cite{Casanueva:2020} has 77 intents about banking, which is challenging due to subtle differences among classes. CLINC \cite{Larson:2019} is a balanced dataset with 150 intents. SNIPS \cite{Coucke:2018} is another balanced dataset covering seven intents. The details of the datasets are given in Table \ref{DataCharacteristics}. We apply no preprocessing.

\begin{table*}[ht]
\centering
\fontsize{10}{11}\selectfont
\caption{Comparison of ConQX with the baselines for intent detection in terms of the weighted F1 score. The means of 10-fold cross-validation are reported. The bold score is the highest. $\bullet$ indicates statistically significant improvement at a 95\% interval in pairwise comparisons between the highest method and baselines marked with $\circ$.} 
\label{ClassificationResults}
\begin{tabular}{lllllll}
\hline
\multirow{2}{*}{\textbf{Expansion Method}} & \multicolumn{2}{c}{\textbf{Banking}} &  \multicolumn{2}{c}{\textbf{CLINC}} & \multicolumn{2}{c}{\textbf{SNIPS}} \\
\cline{2-7}
& {\small BERT} & {\small RoBERTa} & {\small BERT} & {\small RoBERTa} & {\small BERT} & {\small RoBERTa} \\
\hline
Without expansion & 0.908$\circ$ & 0.923 & 0.959 & 0.964 & 0.974 & 0.979 \\
Bag-of-words (kNN) & 0.909$\circ$ & 0.922 & 0.953$\circ$ & 0.957$\circ$ & 0.969$\circ$ & 0.972$\circ$ \\
Transformer (GPT-2) & 0.912 & 0.923 & 0.954 & 0.964 & 0.976 & 0.981 \\
\hline
ConQX (zero-shot) & \textbf{0.920}$\bullet$ & \textbf{0.928} & 0.960 & \textbf{0.965}$\bullet$ & 0.978 & \textbf{0.983}$\bullet$ \\

ConQX (one-shot) & \textbf{0.920}$\bullet$ & \textbf{0.928} & 0.959 & 0.961 & \textbf{0.983}$\bullet$ & 0.981 \\

ConQX (few-shot) & 0.916 & 0.925 & \textbf{0.962}$\bullet$ & 0.962 & 0.981 & 0.976 \\

\hline
\end{tabular}
\end{table*}

\subsection{Experimental Design}

Query expansion is conducted on NVIDIA 2080Ti GPU with 12 GB memory; BERT fine-tuning uses the same infrastructure as well. Query expansion takes approximately an hour to complete in the 10-fold setting. The details of the compared methods are given as follows.

\begin{itemize}[leftmargin=*,noitemsep]
    \item \textbf{Without expansion:} As a baseline method, we fine-tune BERT-base \cite{Devlin:2018} and RoBERTa-base \cite{Liu:2019b} with default parameters by Huggingface \cite{Wolf:2019}, but without expansion. 
    \item \textbf{Bag-of-words (kNN):} As a baseline expansion method, we consider that the expanded words are independent (bag-of-words), and use GloVe \cite{Pennington:2014} word embeddings. We sample \emph{k}=1 nearest neighbor of each input token in the embedding space, and append them to the input, using scikit-learn \cite{Pedregosa:2011}. 
    \item \textbf{Transformer (GPT-2):} As a baseline expansion method, GPT2-large with 774M parameters by Huggingface \cite{Wolf:2019} is used for text generation, given the original query with no input prompt \cite{Radford:2019}. The number of generated tokens is approximated to the number of input tokens.
    \item \textbf{ConQX (zero-shot):} Our semantic expansion method with zero-shot learning. Both train and test instances are expanded, and fine-tuning is done using the expanded queries. For top-k sampling in GPT-2, we experiment $k\in(10, 50, 100)$, and select empirically by F1 score on the test set.  
    \item \textbf{ConQX (one-shot):} Our method with one-shot learning. A single true demonstration of semantic expansion is provided. 
    \item \textbf{ConQX (few-shot):} Our method with few-shot learning. Multiple true demonstrations of semantic expansion are provided (we use four demonstrations for the sake of efficiency). 
\end{itemize}

We report the weighted average F1 score for intent detection with leave-one-out 10-fold cross validation. We use scikit-learn \cite{Pedregosa:2011} for evaluation metrics. Any improvements over the baselines are statistically validated by the two-tailed paired t-test at a 95\% interval.

\subsection{Experimental Results}
We compare the effectiveness of baselines and our method for intent detection in Table \ref{ClassificationResults}. The results show that ConQX improves the effectiveness of intent detection in all datasets, compared to all baselines. Although, the gap between baselines and ConQX is not too wide, we show that the differences are statistically significant in some cases. We show that conditioned text generation is a promising approach for semantic expansion of spoken queries, and its performance can be improved by additional \emph{prompt mining}. ConQX with zero/one-shot learning lead to better improvement in most cases, showing that hand-crafted true demonstrations could cause noise in few-shot learning, i.e. prompt mining can generate better demonstrations. 
GPT-2 without conditioned text generation does not always improve effectiveness, showing the need of conditioned text generation. kNN-based expansion also deteriorates effectiveness in some cases, possibly due to the fact that the neighbor words do not clarify the context of short queries. 

We analyze prompts differing in length, formality of language, and syntactic structure using punctuation. Longer prompts tend to result in more coherent and informative expansions. However, depending on the dataset and classifier, shorter prompts may be more suitable. The role of different prompts are application-dependent, formally written prompts are favored by formal domains such as banking. Syntactic structures, such as quotation marks, make irrelevant text filtered out and result in less noisy expansions.

\section{Conclusion and Future Work}
We propose conditioned query expansion using Transformer-based language models. We show that the performance is increased in all datasets from different domains, with proper selection of parameters (zero/one/few-shot and prompts). ConQX is thereby a promising method for similar tasks that can benefit from semantic expansion.
 
Future work would have more observations on other models and sampling strategies, such as beam search \cite{Shao:2017}. We improve the performance with hand-crafted prompts, but \emph{prompt mining} and \emph{promp engineering} are novel research areas. We plan to focus on tuning parameters, such as zero/one/few-shot learning and a systematic way to generate prompts \cite{jiang-etal-2020-know}. 

\bibliographystyle{acl_natbib}
\bibliography{anthology,emnlp2021}

\begin{thebibliography}{23}
\expandafter\ifx\csname natexlab\endcsname\relax\def\natexlab#1{#1}\fi

\bibitem[{Bengio et~al.(2003)Bengio, Ducharme, Vincent, and
  Janvin}]{Bengio:2003}
Yoshua Bengio, R\'{e}jean Ducharme, Pascal Vincent, and Christian Janvin. 2003.
\newblock A neural probabilistic language model.
\newblock \emph{Journal of Machine Learning Research}, 3:1137–1155.

\bibitem[{Brown et~al.(2020)}]{Brown:2020}
Tom~B Brown et~al. 2020.
\newblock Language models are few-shot learners.
\newblock \emph{arXiv preprint arXiv:2005.14165}.

\bibitem[{Casanueva et~al.(2020)Casanueva, Tem{\v{c}}inas, Gerz, Henderson, and
  Vuli{\'c}}]{Casanueva:2020}
I{\~n}igo Casanueva, Tadas Tem{\v{c}}inas, Daniela Gerz, Matthew Henderson, and
  Ivan Vuli{\'c}. 2020.
\newblock Efficient intent detection with dual sentence encoders.
\newblock \emph{arXiv preprint arXiv:2003.04807}.

\bibitem[{Claveau(2020)}]{Claveau:2020}
Vincent Claveau. 2020.
\newblock Query expansion with artificially generated texts.
\newblock \emph{arXiv preprint arXiv:2012.08787}.

\bibitem[{Coucke et~al.(2018)}]{Coucke:2018}
Alice Coucke et~al. 2018.
\newblock Snips voice platform: {A}n embedded spoken language understanding
  system for private-by-design voice interfaces.
\newblock \emph{arXiv preprint arXiv:1805.10190}.

\bibitem[{Dai and Callan(2019)}]{Dai:2019}
Zhuyun Dai and Jamie Callan. 2019.
\newblock Deeper text understanding for ir with contextual neural language
  modeling.
\newblock In \emph{Proceedings of the 42nd International ACM SIGIR Conference
  on Research and Development in Information Retrieval}, pages 985--988.

\bibitem[{Devlin et~al.(2019)Devlin, Chang, Lee, and Toutanova}]{Devlin:2018}
Jacob Devlin, Ming-Wei Chang, Kenton Lee, and Kristina Toutanova. 2019.
\newblock {BERT}: Pre-training of deep bidirectional transformers for language
  understanding.
\newblock In \emph{Proceedings of the 2019 Conference of the North American
  Chapter of the Association for Computational Linguistics: Human Language
  Technologies}, pages 4171--4186.

\bibitem[{Fan et~al.(2018)Fan, Lewis, and Dauphin}]{Fan:2018}
Angela Fan, Mike Lewis, and Yann Dauphin. 2018.
\newblock Hierarchical neural story generation.
\newblock In \emph{Proceedings of the 56th Annual Meeting of the Association
  for Computational Linguistics (Volume 1: Long Papers)}, pages 889--898.

\bibitem[{Gao et~al.(2020)Gao, Fisch, and Chen}]{Gao:2020}
Tianyu Gao, Adam Fisch, and Danqi Chen. 2020.
\newblock Making pre-trained language models better few-shot learners.
\newblock \emph{arXiv preprint arXiv:2012.15723}.

\bibitem[{Jiang et~al.(2020)Jiang, Xu, Araki, and
  Neubig}]{jiang-etal-2020-know}
Zhengbao Jiang, Frank~F. Xu, Jun Araki, and Graham Neubig. 2020.
\newblock \href {https://doi.org/10.1162/tacl_a_00324} {How can we know what
  language models know?}
\newblock \emph{Transactions of the Association for Computational Linguistics},
  8:423--438.

\bibitem[{Larson et~al.(2019)}]{Larson:2019}
Stefan Larson et~al. 2019.
\newblock An evaluation dataset for intent classification and out-of-scope
  prediction.
\newblock In \emph{Proceedings of the 2019 Conference on Empirical Methods in
  Natural Language Processing and the 9th International Joint Conference on
  Natural Language Processing (EMNLP-IJCNLP)}, pages 1311--1316.

\bibitem[{Liu et~al.(2019)Liu, Ott, Goyal, Du, Joshi, Chen, Levy, Lewis,
  Zettlemoyer, and Stoyanov}]{Liu:2019b}
Yinhan Liu, Myle Ott, Naman Goyal, Jingfei Du, Mandar Joshi, Danqi Chen, Omer
  Levy, Mike Lewis, Luke Zettlemoyer, and Veselin Stoyanov. 2019.
\newblock Roberta: A robustly optimized bert pretraining approach.
\newblock \emph{arXiv preprint arXiv:1907.11692}.

\bibitem[{Padaki et~al.(2020)Padaki, Dai, and Callan}]{Padaki:2020}
Ramith Padaki, Zhuyun Dai, and Jamie Callan. 2020.
\newblock Rethinking query expansion for {BERT} reranking.
\newblock In \emph{42nd European Conference on {IR} Research, {ECIR}}, pages
  297--304.

\bibitem[{Pedregosa et~al.(2011)}]{Pedregosa:2011}
F.~Pedregosa et~al. 2011.
\newblock Scikit-learn: Machine learning in {P}ython.
\newblock \emph{Journal of Machine Learning Research}, 12:2825--2830.

\bibitem[{Pennington et~al.(2014)Pennington, Socher, and
  Manning}]{Pennington:2014}
Jeffrey Pennington, Richard Socher, and Christopher~D Manning. 2014.
\newblock Glove: Global vectors for word representation.
\newblock In \emph{Proceedings of the 2014 Conference on Empirical Methods in
  Natural Language Processing (EMNLP)}, pages 1532--1543.

\bibitem[{Radford et~al.(2019)}]{Radford:2019}
Alec Radford et~al. 2019.
\newblock Language models are unsupervised multitask learners.
\newblock \emph{OpenAI Blog}, 1(8):9.

\bibitem[{Roy et~al.(2016)Roy, Paul, Mitra, and Garain}]{Roy:2016}
Dwaipayan Roy, Debjyoti Paul, Mandar Mitra, and Utpal Garain. 2016.
\newblock Using word embeddings for automatic query expansion.
\newblock \emph{arXiv preprint arXiv:1606.07608}.

\bibitem[{Schick and Schütze(2020)}]{Schick:2020}
Timo Schick and Hinrich Schütze. 2020.
\newblock It's not just size that matters: Small language models are also
  few-shot learners.
\newblock \emph{arXiv preprint arXiv:2009.07118}.

\bibitem[{Shao et~al.(2017{\natexlab{a}})Shao, Gouws, Britz, Goldie, Strope,
  and Kurzweil}]{shao-etal-2017-generating}
Yuanlong Shao, Stephan Gouws, Denny Britz, Anna Goldie, Brian Strope, and Ray
  Kurzweil. 2017{\natexlab{a}}.
\newblock \href {https://doi.org/10.18653/v1/D17-1235} {Generating high-quality
  and informative conversation responses with sequence-to-sequence models}.
\newblock In \emph{Proceedings of the 2017 Conference on Empirical Methods in
  Natural Language Processing}, pages 2210--2219, Copenhagen, Denmark.
  Association for Computational Linguistics.

\bibitem[{Shao et~al.(2017{\natexlab{b}})}]{Shao:2017}
Yuanlong Shao et~al. 2017{\natexlab{b}}.
\newblock Generating high-quality and informative conversation responses with
  sequence-to-sequence models.
\newblock In \emph{Proceedings of the 2017 Conference on Empirical Methods in
  Natural Language Processing}, pages 2210--2219.

\bibitem[{{Tur} et~al.(2010){Tur}, {Hakkani-Tür}, and {Heck}}]{Tur:2010}
G.~{Tur}, D.~{Hakkani-Tür}, and L.~{Heck}. 2010.
\newblock What is left to be understood in {ATIS}?
\newblock In \emph{2010 IEEE Spoken Language Technology Workshop}, pages
  19--24.

\bibitem[{Vaswani et~al.(2017)}]{Vaswani:2017}
Ashish Vaswani et~al. 2017.
\newblock Attention is all you need.
\newblock In \emph{Proceedings of the 31st International Conference on Neural
  Information Processing Systems}, pages 6000–--6010.

\bibitem[{Wolf et~al.(2019)}]{Wolf:2019}
Thomas Wolf et~al. 2019.
\newblock Huggingface's transformers: State-of-the-art natural language
  processing.
\newblock \emph{arXiv preprint arXiv:1910.03771}.

\end{thebibliography}


\end{document}